%% file: PaperForReview.tex
\documentclass[10pt,twocolumn,letterpaper]{article}

\usepackage[pagenumbers]{cvpr}      
\usepackage{ulem}
\usepackage{graphicx}
\usepackage{amsmath}
\usepackage{amssymb}
\usepackage{booktabs}
\usepackage{enumitem}
\usepackage{tabulary}
\usepackage{adjustbox}
\usepackage{silence}
\usepackage{rotating}
\usepackage{lipsum}

\usepackage{caption}
\usepackage{array}

\usepackage[pagebackref,breaklinks,colorlinks]{hyperref}

\usepackage[capitalize]{cleveref}
\crefname{section}{Sec.}{Secs.}
\Crefname{section}{Section}{Sections}
\Crefname{table}{Table}{Tables}
\crefname{table}{Tab.}{Tabs.}

\newcommand{\tocite}[1]{\textcolor{red}{[TOCITE]}}

\def\confName{CVPR}
\def\confYear{2023}

\begin{document}

\title{Painting 3D Nature in 2D: \\
View Synthesis of Natural Scenes from a Single Semantic Mask}
\author{
Shangzan Zhang$^{1}$
\quad
Sida Peng$^{1}$
\quad
Tianrun Chen$^{1}$
\quad
Linzhan Mou$^1$ 
\quad
Haotong Lin$^1$\\[1.5mm]
\quad
Kaicheng Yu$^2$
\quad
Yiyi Liao$^1$
\quad
Xiaowei Zhou$^1$
\\[1.5mm]
$^1$Zhejiang University
\quad
$^2$Alibaba Group
}

\twocolumn[{%
\renewcommand\twocolumn[1][]{#1}%
\maketitle
\begin{center}
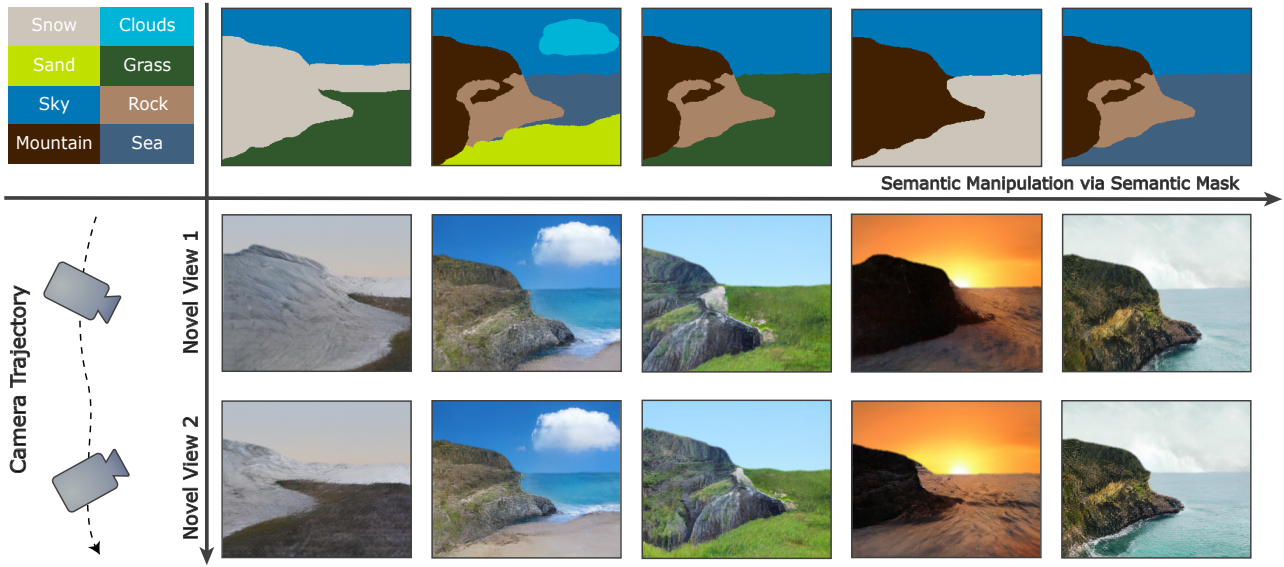

\captionsetup{type=figure}

\input{fig/teaser.tex}

\captionsetup{font={normalsize}} 
\captionof{figure}{Given only a single semantic map as input (first row), our approach optimizes neural fields for view synthesis of natural scenes.
Photorealistic images can be rendered via neural fields (the last two rows).}
\vspace{+0.2in}

\label{fig:teaser}
\end{center}%
}]


\input{shortcuts}

\input{sec_0_abstract.tex}

\input{sec_1_intro.tex}
\input{sec_2_related_v2.tex}
\input{sec_3_method_v3.tex}
\input{sec_4_exp.tex}
\input{sec_5_ablation.tex}
\input{sec_6_conclusion.tex}

{\small
\bibliographystyle{ieee_fullname}
\bibliography{egbib}
}
\newpage
\;
\input{Supp_arxiv.tex}

\end{document}

%% file: fig/teaser.tex
\vspace{-2em}
\centering
\includegraphics[width=1\linewidth]{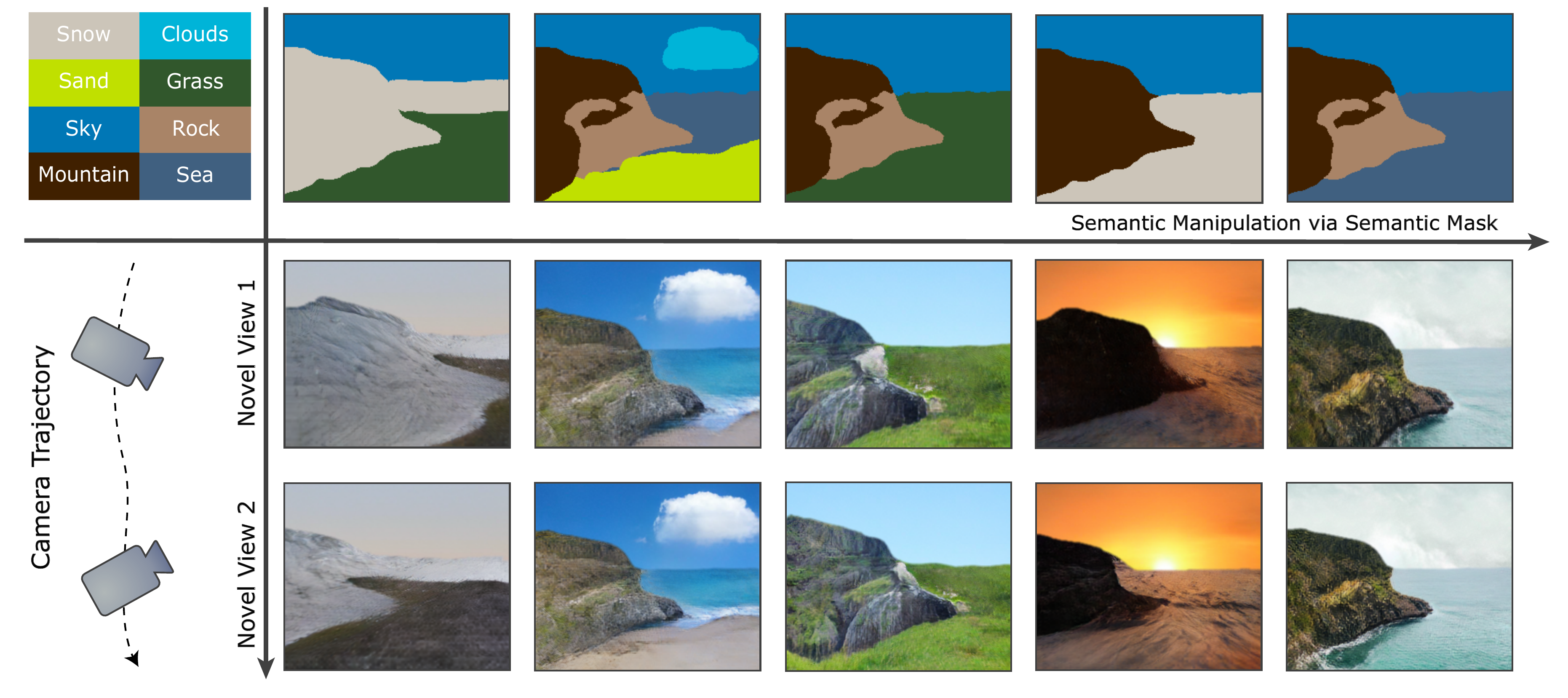}
\vspace{-2em}

%% file: shortcuts.tex
\newcommand{\Perp}{\perp\!\!\! \perp}
\newcommand{\bK}{\mathbf{K}}
\newcommand{\bX}{\mathbf{X}}
\newcommand{\bY}{\mathbf{Y}}
\newcommand{\bk}{\mathbf{k}}
\newcommand{\bx}{\mathbf{x}}
\newcommand{\by}{\mathbf{y}}
\newcommand{\bhy}{\hat{\mathbf{y}}}
\newcommand{\bty}{\tilde{\mathbf{y}}}
\newcommand{\bG}{\mathbf{G}}
\newcommand{\bI}{\mathbf{I}}
\newcommand{\bg}{\mathbf{g}}
\newcommand{\bS}{\mathbf{S}}
\newcommand{\bs}{\mathbf{s}}
\newcommand{\bM}{\mathbf{M}}
\newcommand{\bw}{\mathbf{w}}
\newcommand{\eye}{\mathbf{I}}
\newcommand{\bU}{\mathbf{U}}
\newcommand{\bV}{\mathbf{V}}
\newcommand{\bW}{\mathbf{W}}
\newcommand{\bn}{\mathbf{n}}
\newcommand{\bv}{\mathbf{v}}
\newcommand{\bq}{\mathbf{q}}
\newcommand{\bR}{\mathbf{R}}
\newcommand{\bi}{\mathbf{i}}
\newcommand{\bj}{\mathbf{j}}
\newcommand{\bp}{\mathbf{p}}
\newcommand{\bt}{\mathbf{t}}
\newcommand{\bJ}{\mathbf{J}}
\newcommand{\bu}{\mathbf{u}}
\newcommand{\bB}{\mathbf{B}}
\newcommand{\bD}{\mathbf{D}}
\newcommand{\bz}{\mathbf{z}}
\newcommand{\bP}{\mathbf{P}}
\newcommand{\bC}{\mathbf{C}}
\newcommand{\bA}{\mathbf{A}}
\newcommand{\bZ}{\mathbf{Z}}
\newcommand{\bff}{\mathbf{f}}
\newcommand{\bF}{\mathbf{F}}
\newcommand{\bo}{\mathbf{o}}
\newcommand{\bO}{\mathbf{O}}
\newcommand{\bc}{\mathbf{c}}
\newcommand{\bm}{\mathbf{m}}
\newcommand{\bT}{\mathbf{T}}
\newcommand{\bQ}{\mathbf{Q}}
\newcommand{\bL}{\mathbf{L}}
\newcommand{\bl}{\mathbf{l}}
\newcommand{\ba}{\mathbf{a}}
\newcommand{\bE}{\mathbf{E}}
\newcommand{\bH}{\mathbf{H}}
\newcommand{\bd}{\mathbf{d}}
\newcommand{\br}{\mathbf{r}}
\newcommand{\be}{\mathbf{e}}
\newcommand{\bb}{\mathbf{b}}
\newcommand{\bh}{\mathbf{h}}
\newcommand{\bhh}{\hat{\mathbf{h}}}
\newcommand{\btheta}{\boldsymbol{\theta}}
\newcommand{\bTheta}{\boldsymbol{\Theta}}
\newcommand{\bpi}{\boldsymbol{\pi}}
\newcommand{\bphi}{\boldsymbol{\phi}}
\newcommand{\bPhi}{\boldsymbol{\Phi}}
\newcommand{\bmu}{\boldsymbol{\mu}}
\newcommand{\bSigma}{\boldsymbol{\Sigma}}
\newcommand{\bGamma}{\boldsymbol{\Gamma}}
\newcommand{\bbeta}{\boldsymbol{\beta}}
\newcommand{\bomega}{\boldsymbol{\omega}}
\newcommand{\blambda}{\boldsymbol{\lambda}}
\newcommand{\bLambda}{\boldsymbol{\Lambda}}
\newcommand{\bkappa}{\boldsymbol{\kappa}}
\newcommand{\btau}{\boldsymbol{\tau}}
\newcommand{\balpha}{\boldsymbol{\alpha}}
\newcommand{\nR}{\mathbb{R}}
\newcommand{\nN}{\mathbb{N}}
\newcommand{\nL}{\mathbb{L}}
\newcommand{\nF}{\mathbb{F}}
\newcommand{\nS}{\mathbb{S}}
\newcommand{\cN}{\mathcal{N}}
\newcommand{\cM}{\mathcal{M}}
\newcommand{\cR}{\mathcal{R}}
\newcommand{\cB}{\mathcal{B}}
\newcommand{\cL}{\mathcal{L}}
\newcommand{\cH}{\mathcal{H}}
\newcommand{\cS}{\mathcal{S}}
\newcommand{\cT}{\mathcal{T}}
\newcommand{\cO}{\mathcal{O}}
\newcommand{\cC}{\mathcal{C}}
\newcommand{\cP}{\mathcal{P}}
\newcommand{\cE}{\mathcal{E}}
\newcommand{\cF}{\mathcal{F}}
\newcommand{\cK}{\mathcal{K}}
\newcommand{\cY}{\mathcal{Y}}
\newcommand{\cX}{\mathcal{X}}
\newcommand{\cV}{\mathcal{V}}
\def\bgamma{\boldsymbol\gamma}

\newcommand{\specialcell}[2][c]{%
  \begin{tabular}[#1]{@{}c@{}}#2\end{tabular}}

\renewcommand{\b}{\ensuremath{\mathbf}}

\def\mc{\mathcal}
\def\mb{\mathbf}

\newcommand{\T}{^{\raisemath{-1pt}{\mathsf{T}}}}

\makeatletter
\DeclareRobustCommand\onedot{\futurelet\@let@token\@onedot}
\def\@onedot{\ifx\@let@token.\else.\null\fi\xspace}
\def\eg{e.g\onedot} \def\Eg{E.g\onedot}
\def\ie{i.e\onedot} \def\Ie{I.e\onedot}
\def\cf{cf\onedot} \def\Cf{Cf\onedot}
\def\etc{etc\onedot} \def\vs{vs\onedot}
\def\wrt{wrt\onedot}
\def\dof{d.o.f\onedot}
\def\etal{\textit{et~al}\onedot} \def\iid{i.i.d\onedot}
\def\Fig{Fig\onedot} \def\Eqn{Eqn\onedot} \def\Sec{Sec\onedot} \def\Alg{Alg\onedot}
\makeatother

\newcommand{\figref}[1]{\Fig~\ref{#1}}
\newcommand{\secref}[1]{Section~\ref{#1}}
\newcommand{\appref}[1]{Appendix~\ref*{#1}}
\newcommand{\algref}[1]{Algorithm~\ref{#1}}
\renewcommand{\eqref}[1]{Eq.~\ref{#1}}
\newcommand{\tabref}[1]{Table~\ref{#1}}

\newcommand\blfootnote[1]{%
\begingroup
\renewcommand\thefootnote{}\footnote{#1}%
\addtocounter{footnote}{-1}%
\endgroup
}

\renewcommand\UrlFont{\color{black}\rmfamily}

\newcommand*\rot{\rotatebox{90}}
\newcommand{\boldparagraph}[1]{\vspace{0.2cm}\noindent{\bf #1:} }


\newif\ifcomment
\commenttrue



%% file: sec_0_abstract.tex
\begin{abstract}
We introduce a novel approach that takes a single semantic mask as input to synthesize multi-view consistent color images of natural scenes, trained with  a collection of single images from the Internet.
Prior works on 3D-aware image synthesis either require multi-view supervision or learning category-level prior for specific classes of objects, which can hardly work for natural scenes. 
Our key idea to solve this challenging problem is to use a semantic field as the intermediate representation, which is easier to reconstruct from an input semantic mask and then translate to a radiance field with the assistance of off-the-shelf semantic image synthesis models. 
Experiments show that our method outperforms baseline methods and produces photorealistic, multi-view consistent videos of a variety of natural scenes. 
Project website: \href{https://zju3dv.github.io/paintingnature/}{https://zju3dv.github.io/paintingnature/}
 \end{abstract}

%% file: sec_1_intro.tex
\section{Introduction}
\label{sec:intro}



%
Natural scenes are indispensable
content in many applications such as film production and
video games.
This work focuses on a specific setting of synthesizing novel views of natural scenes given a single semantic mask, which enables us to generate 3D contents by editing 2D semantic masks.
With the development of deep generative models, 2D semantic image synthesis methods~\cite{Park2019SemanticIS, sushko2020you, wang2018high, isola2017image} have achieved impressive advances.
However, they do not consider the underlying 3D world and cannot generate multi-view consistent free-viewpoint videos.


To address this problem, a straightforward approach is first utilizing semantics-driven image generator like SPADE~\cite{Park2019SemanticIS} to synthesize an image from the input semantic mask and then predicting novel views based on the generated image.
Although the existing single-view view synthesis methods~\cite{Liu2021InfiniteNP,yu2021pixelnerf, li2021mine, Park2022BridgingIA, Wiles2020SynSinEV,Rombach2021GeometryFreeVS} achieve impressive rendering results, they typically require training networks on posed multi-view images. Compared to urban or indoor scenes, learning
to synthesize natural scenes is a challenging
task as it is difficult to collect 3D data or posed videos of
natural scenes for training, as demonstrated in~\cite{Li2022InfiniteNatureZeroLP}, making the
aforementioned methods not applicable.
AdaMPI~\cite{han2022single} designs a training strategy to learn the view synthesis network on single-view image collections.
It warps images to random novel views and warps them back to the original view.
An inpainting network is trained to fill the holes in disocclusion regions to match the original images.
After training, the inpainting network is used to generate pseudo multi-view images for training a view synthesis network.
Our experimental results in Section \ref{ablation} show that the inpainting network struggles to output high-quality image contents in missing regions under large viewpoint changes, thus limiting the rendering quality.


In this paper, we propose a novel framework for semantics-guided view synthesis of natural scenes by learning prior from single-view image collections.
Based on the observation that semantic masks have much lower complexity than images, we exploratively divide this task into two simpler subproblems: we first generate semantic masks at novel views and then translate them to RGB images through SPADE.
For view synthesis of semantic masks, the input semantic mask is first translated to a color image by SPADE, and a depth map is predicted from the color image by a depth estimator~\cite{ranftl2020towards}.
Then the input semantic mask is warped to novel views using the predicted depth map and refined by an inpainting network, trained by a self-supervised learning strategy on single-view image collections.
Our experiments show that, in contrast to images, the novel view synthesis of semantic masks is much easier to learn by the network.


It is observed that semantic masks generated by the inpainting network tend to be view inconsistent.
Although the inconsistency across views seems minor, SPADE could generate quite different contents in these regions.
\figref{fig:vis_fuse} presents two examples.
To solve this issue, we learn a neural semantic field to fuse and denoise these semantic masks for better multi-view consistency.
Finally, we translate the multi-view semantic masks to color images by SPADE and reconstruct a neural scene representation for view-consistent rendering.

Extensive experiments are conducted on the LHQ dataset~\cite{skorokhodov2021aligning}, a widely-used benchmark dataset for semantic image synthesis.
The results demonstrate that our approach significantly outperforms baseline methods both qualitatively and quantitatively.
We also show that by editing the input semantic mask, our approach is capable of generating various high-quality rendering results of natural scenes, as shown in \figref{fig:teaser}.



%% file: sec_2_related_v2.tex

\section{Related Works}
\label{sec:formatting}

\paragraph{Semantics-guided view synthesis.}
This task takes a single 2D semantic mask as input and outputs free-viewpoint videos of the 3D scene. Only a few works attempt to tackle this challenging task, including  SVS~\cite{Huang2020SemanticVS} and GVS~\cite{Habtegebrial2020GenerativeVS}. Given posed videos, 
SVS~\cite{Huang2020SemanticVS} first extracts the semantic map for each video frame. During training, it leverages a SPADE-like generator~\cite{Park2019SemanticIS} to produce an RGB image from the input semantic mask at a particular frame and then regresses from the image to the multi-plane images (MPI) as the generated 3D scene for view synthesis. The generated MPI is rendered into several viewpoints, and the SVS model is optimized by minimizing the difference between rendered images and observed images of the input video.  However, they train their models on datasets of posed videos ~\cite{Zhou2018StereoML, cabon2020virtual,dosovitskiy2017carla}. A problem with such a training strategy is that a large number of videos with calibrated camera poses are hard to obtain in real life, which could be expensive and limit the diversity of training data. Qiao et al.~\cite{qiao2022learning} mainly focus on novel-view scene layout generation and their model is trained on posed images, while we concentrate on rendering consistent RGB images with single image collections
for training.
 

\paragraph{Single-view view synthesis.}
 Recently, many works have focused on single-view view synthesis with monocular RGB image as the input to generate free-viewpoint videos~\cite{kopf2020one, hu2021worldsheet, Wiles2020SynSinEV, Shih20203DPU,Liu2021InfiniteNP,Niklaus20193DKB,Tulsiani2018Layerstructured3S,Rockwell2021PixelSynthGA,Rombach2021GeometryFreeVS}. Among existing works, some use explicit 3D representations, such as 
layered depth image (LDI)~\cite{Hedman2018Instant3P, shade1998layered} and multi-plane image (MPI)~\cite{Zhou2018StereoML}, which can capture visible contents and infer disocclusion regions. Another line of research predicts neural radiance fields~\cite{mildenhall2021nerf} from a single image. For example, PixelNeRF~\cite{yu2021pixelnerf} extracts image features using
2D CNN and constructs an aligned feature field to render novel views. 
Li \etal ~\cite{li2021mine} combine MPI and NeRF representation and learn a continuous 3D field. These works learn 3D representations to perform novel-view synthesis, and the learning is mainly based on multi-view images or posed videos for supervision. 
However, similar to the dataset constraint in semantic view synthesis, large-scale multi-view datasets are also rare, thus bringing challenges for high-quality view synthesis for natural scenes. 

Recently, some methods have utilized single-view image collections to train a neural network that performs novel view synthesis given a single-view image. For example,  
~\cite{kopf2020one} and~\cite{Shih20203DPU} use a monocular depth estimation network to construct LDI representation and leverage the inpainting network to synthesize disocclusion content to perform view synthesis. 
AdaMPI~\cite{han2022single} proposes a warp-back training strategy to train MPI on the COCO~\cite{caesar2018coco} datasets.
Li \etal~\cite{Li2022InfiniteNatureZeroLP} propose a cycle-rendering strategy that warps the input image to a virtual view and then warps it back to the original view, finally resulting in an image containing holes.
A refinement network is trained to refine the final image to match the original image.
They tailored a viable path by demonstrating the capability of single-view image collections collected from the Internet in monocular view synthesis. 
Still, no prior works have attempted to train models solely
using single-view image collections while performing novel view synthesis from a single semantic mask.

A naive solution to leverage existing single-view image collections~\cite{skorokhodov2021aligning} to learn models for semantics-guided semantic view synthesis is extending the above approaches
~\cite{Shih20203DPU, han2022single} 
through a two-stage scheme: first converting the semantic mask to an RGB image and then applying the view synthesis model.
However, this naive solution does not fully utilize semantic information, thus leading to poor performance.

\paragraph{Neural Radiance Fields.}
%
Neural Radiance Fields~\cite{mildenhall2021nerf} and its subsequent works significantly advance the realm of novel view synthesis~\cite{zhang2020nerf++, barron2022mip, barron2021mip,niemeyer2022regnerf, xu2022sinnerf} and 3D reconstruction~\cite{oechsle2021unisurf,zhang2022nerfusion,ehret2022nerf}.
While the above works focus on rendering realistic novel view images or reconstructing accurate 3D geometry, 
some works currently aim to exploit 2D pretrained models to learn a priori knowledge for the neural field.
Examples include creating 3D objects driven by text~\cite{wang2022clip,jain2022zero,poole2022dreamfusion}, animating  NeRF by audio signals~\cite{guo2021ad, liu2022semantic}, and stylizing scenes~\cite{zhang2022arf, huang2022stylizednerf,nguyen2022snerf}.
Unlike the previous works, in
this paper, we focus on utilizing semantic image synthesis models.
Semantic-NeRF~\cite{zhi2021place} and later works~\cite{fu2022panoptic, kundu2022panoptic,tschernezki2022neural} use NeRF as a powerful 3D fusion tool to fuse 2D semantic information. 
However, they do not explore how to recover a semantic field from only a single semantic mask.
Another line of work~\cite{chen2022sem2nerf,sun2022fenerf,sun2022ide} extends 3D-aware generative modeling~\cite{schwarz2020graf, schwarz2022voxgraf,niemeyer2021giraffe,chan2021pi,niemeyer2021campari}
to edit 3D appearance and geometry via a semantic mask, but they mainly conduct their experiments on object-centric datasets (e.g., FFHQ~\cite{kazemi2014one}) with known camera distribution and fail to generate complex scenes on the nature scene datasets.
A closely related work GANcraft~\cite{hao2021gancraft} and its extended work~\cite{jeong20223d} use image-to-image translation techniques~\cite{Park2019SemanticIS,sushko2020you,wang2018high,isola2017image} to synthesize pseudo ground truths and discriminators to make generative free-viewpoint videos more realistic from 3D semantic labels. Nevertheless, they need 3D semantic labels to render consistent 2D semantic masks and fail when they only take a single semantic mask as input.  

\paragraph{Image-to-image translations}
Image-to-image translations~\cite{Park2019SemanticIS,sushko2020you,wang2018high,isola2017image} have made tremendous development and can synthesize realistic natural images. 
Pix2Pix~\cite{isola2017image} firstly leverages conditional GAN~\cite{mirza2014conditional} to improve the performance of semantic image synthesis. The performance of image synthesis is significantly enhanced
by SPADE~\cite{Park2019SemanticIS}, which proposes a spatial-varying normalization layer.
OASIS~\cite{sushko2020you} proposes a novel discriminator for semantic image synthesis~\cite{johnson2016perceptual}. However, these works can only synthesize 2D images and are not able to generate images of different views. 


%% file: sec_3_method_v3.tex
\section{Method}
\begin{figure*}[t]
\centering
\includegraphics[width=1\linewidth]{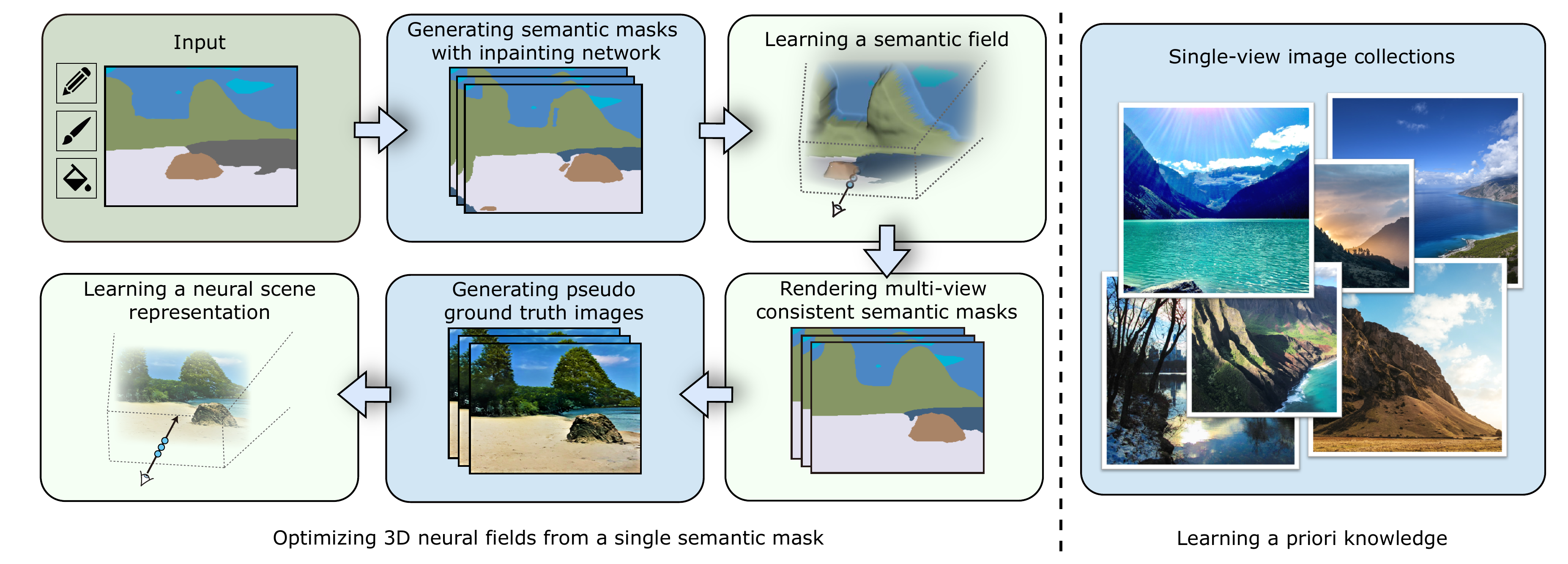}
\vspace{-1em}
\captionsetup{font={normalsize}} 
\caption{\textbf{Illustration of our pipeline.}
\textbf{Left}: Our pipeline can be divided into two steps: we first generate multi-view semantic masks with an inpainting network and then convert semantic masks to RGB images using SPADE.
In order to denoise and fuse semantic information, a semantic field is learned for rendering multi-view consistent masks. 
Finally, a neural scene representation is optimized to fuse appearance information provided by SPADE, which enables view-consistent rendering.
\textbf{Right}: Our semantic inpainting network and SPADE are trained on single-view image collections.}
\label{fig:pipeline}
\end{figure*}
Our goal is to perform photorealistic view synthesis of natural scenes, given a single semantic mask, by learning prior from single-view image collections.
To this end, we divide this task into two simpler subproblems, as described in this section.
%
We first introduce how to generate multi-view consistent semantic masks from the given semantic mask (Section~\ref{ssnerf}).
%
Then, Section~\ref{sgr} discusses how to translate the multi-view semantic masks to color images by SPADE and recover a neural scene representation for view-consistent rendering.
%
%
%
%

\subsection{Generating view-consistent semantic masks}\label{ssnerf}

Fig.~\ref{fig:pipeline} illustrates the overview of generating multi-view consistent semantic masks from a single semantic mask.
%
%
Specifically, we first warp the given semantic mask to novel views.
Then, an inpainting network is utilized to fill in the disocclusion areas of the warped semantic mask at each novel view.
After obtaining multiple infilled semantic masks at different viewpoints, we recover a neural semantic field that can fuse and denoise the multi-view semantic information. Finally, multi-view semantic masks can be obtained by the semantic field.

\paragraph{Warping semantic mask.}\label{sw}
%
Our approach warps the given semantic map to novel views through the depth-based warping technique.
We first convert the input semantic map to the corresponding RGB image using SPADE~\cite{Park2019SemanticIS}, and then use a monocular depth estimation network~\cite{ranftl2020towards} to predict the depth map from the generated RGB image.
Then, a 3D triangular mesh is constructed based on the predicted depth following Shih \etal \cite{Shih20203DPU}.
The semantic mask is lifted to the mesh, whose vertices' color is assigned as the corresponding semantic label.
%
The generated 3D triangular mesh in such a manner may contain spurious edges due to the depth discontinuities in the depth map.
To solve this problem, we remove the edges whose vertices are far from each other.  
Eventually, we warp the semantic mask to the novel views using a mesh renderer following~\cite{Shih20203DPU}.
%

\paragraph{Semantic mask inpainting.}\label{sr}
\input{fig/warpback}

Directly warping the given semantic mask to novel views brings many holes in disocclusion regions.
To inpaint the missing contents, we train a semantic inpainting network on single-view natural image collections~\cite{skorokhodov2021aligning} using the self-supervised technique in~\cite{han2022single}.
\figref{fig:warpback} shows our training strategy for semantic inpainting networks.
Specifically, we first use a pre-trained image segmentation model~\cite{chen2017deeplab} and
a monocular depth estimation model~\cite{ranftl2020towards} to produce semantic masks and depth maps for the natural images.
At each training iteration, an image is randomly sampled from the dataset as the source image $\mathbf{I}_i$.
We then use the corresponding depth map $\mathbf{\hat{D}}_i$ to warp the original semantic mask $\mathbf{S}_i$ to a random target view $j$, producing warped semantic mask $\mathbf{S}_{i\rightarrow j}$ and depth map $\mathbf{\hat{D}}_{i\rightarrow j}$ at target view $j$.
Next, we warp semantic mask $\mathbf{S}_{i\rightarrow j}$ back to the source view using depth map $\mathbf{\hat{D}}_{i\rightarrow j}$, which generates a semantic mask $\mathbf{S}_{i\rightarrow j\rightarrow i}$ with holes. 
Finally, we input $\mathbf{S}_{i\rightarrow j\rightarrow i}$ into the semantic inpainting network and train the network to infill these holes, which is supervised with the original semantic mask $\mathbf{S}_i$.


At test time, we first randomly sample a set of viewpoints, then warp the given semantic mask $\mathbf{S}_0$ to these viewpoints to generate warped semantic masks, and finally apply the inpainting network to fill in their disocclusion regions to generate the infilled multi-view semantic masks.

\paragraph{Semantic field fusion.}
\label{sff}
\input{fig/vis_fuse}

%
We observed that the infilled semantic masks are not view-consistent.
Although the artifact regions in semantic masks seem trivial, the generated images from SPADE could be very different on these regions across different viewpoints.
\figref{fig:vis_fuse} presents an example.
To tackle this problem, a semantic field is introduced to fuse and denoise infilled semantic masks.
%
%
%
We adopt a continuous neural field to represent the semantics and geometry of a 3D scene, similar to~\cite{guo2022manhattan}.
For any query point $\mathbf{x}$ in 3D space, an MLP network $f_\theta$ maps it to an SDF value $d$ and an intermediate feature $\mathbf{z}$, and another MLP network $f_\phi$ maps $\mathbf{z}$ to a semantic logits $\mathbf{s}$.
The neural field is defined as:
%
%
\begin{equation}\label{sn}
\begin{aligned}
    &f_\theta: \mathbf{x} \in \mathbb{R}^3 \mapsto\left(d \in \mathbb{R}, \mathbf{z} \in \mathbb{R}^c\right) \\
    &f_\phi: \mathbf{z} \in \mathbb{R}^c \mapsto \mathrm{\mathbf{s}} \in \mathbb{R}^{M_s},
\end{aligned}
\end{equation}
where $M_s$ denotes the number of semantic classes. 
We render the semantic field into semantic logits and depth through the SDF-based volume rendering~\cite{yariv2021volume,wang2021neus}.
%

Considering that the sky is very distant from the foreground, we handle the foreground and the sky separately, following the practice in~\cite{hao2021gancraft}.
The sky is assumed to be a distant 2D plane, and the semantic probability of the sky is defined as a constant one-hot vector $\mathbf{P}_{\text{sky}}$.
The final semantic probability is formulated as:
\begin{equation}
    \mathbf{Y}(\mathbf{r}) = \mathbf{P}_{\text{fg}}(\mathbf{r}) T_{\text{fg}}(\mathbf{r})+ (1 - T_{\text{fg}}(\mathbf{r}))\mathbf{P}_{\text{sky}},
\end{equation}
where $T_{\text{fg}}$ is the accumulated transmittance of the foreground along camera ray $\mathbf{r}$, and $\mathbf{P}_{\text{fg}}$ is the semantic probability obtained by applying the softmax layer to the rendered semantic logits.

%

%
To learn the semantic field, the cross entropy loss is applied to compare the rendered semantic probability $\mathbf{P(\mathbf{r})}$ and the semantic probability $\mathbf{P}^*(\mathbf{r})$ provided by the infilled semantic masks:

\begin{equation}\label{sloss}
\mathcal{L}_{\mathbf{P}}=-\sum_{\mathbf{r} \in \mathcal{R}} \sum_{k=1}^{M_s} \mathbf{P}_k^*(\mathbf{r}) \log \mathbf{P}_k(\mathbf{r}).
\end{equation}

%

In addition, we use depth maps to learn the geometry of the semantic field.
%
%
%
In detail, the infilled semantic masks are first converted to RGB images via SPADE, then processed by a monocular depth estimation network~\cite{ranftl2020towards} to predict depth maps.
A scale- and shift-invariant loss \cite{ranftl2020towards, yu2022monosdf} is utilized to calculate the difference between the rendered depth map 
$\mathbf{D}$ and the predicted depth map $\mathbf{\hat{D}}$, which is defined as:
\begin{equation}
\mathcal{L}_{\text{depth}}=\sum_{\mathbf{r} \in \mathcal{R}'}\|(w \mathbf{D}(\mathbf{r})+q)-\mathbf{\hat{D}}(\mathbf{r})\|^2,
\end{equation}
where $\mathcal{R}'$ means camera rays of image pixels excluding the sky region. $w$ and $q$ are used to align the scale and shift of $\mathbf{D}$ and $\mathbf{\hat{D}}$, which can be obtained using a least-squares criterion~\cite{ranftl2020towards,eigen2014depth}.

To separate the foreground and the sky region, we apply a loss on the accumulated transmittance:
\begin{equation}
    \mathcal{L}_{\text{trans}} = \sum_{\mathbf{r} \in \mathcal{R}} (\log(T_\text{fg}(\mathbf{r}))+\log(1-T_\text{fg}(\mathbf{r}))). 
\end{equation}
%
This loss enforces the transmittance to be either 0 or 1.
The overall loss is described in the supplementary material.

\subsection{Natural scene representations}
\label{sgr}
Directly translating multi-view semantic masks obtained in Section~\ref{sff} to RGB images through SPADE fails to produce multi-view consistent rendering, as shown in the supplementary material. To resolve this issue, we learn a natural scene representation to fuse appearance information provided by SPADE.
This section describes the generation of natural scenes from the learned semantic field.
%
We first introduce the neural representation of natural scenes.
Then, the rendering and training of the scene representation are described.
%

%
The geometry of the scene is directly modeled as the trained MLP network $f_{\theta}$ (Eq.~\ref{sn}) of the semantic field. To represent the scene's appearance, we recover an appearance field $f_{\xi}$. 
Following EG3D~\cite{chan2022efficient}, a tri-plane feature map is adopted to map a point to a feature vector. Specifically, given a point $\mathbf{x}$, it is orthogonally projected to the feature planes to retrieve three feature vectors, which are concatenated into the final feature vector.
We use an MLP network to regress the RGB value $\bc$ from the aggregated feature vector. The appearance field is defined as:
\begin{equation}\label{af}
    f_\xi: \mathbf{x} \in \mathbb{R}^3 \mapsto\bc\in\nR^3
\end{equation}

For the scene representation, we separately model the foreground and the sky region.
The sky is implemented as a 2D image plane generated by a 2D generator network, and we place it at a distance.
%
%
For a ray classified as ‘sky’, the sky image plane maps the intersection point $(u,v)$ between the ray and the sky plane to the RGB value.

To efficiently render the scene representation, we leverage the pre-learned scene geometry to guide the sampling of points along camera rays.
%
The mesh is first extracted from the trained MLP network $f_{\theta}$. 
Then, for each camera ray, we only predict the color for the point on the mesh surface, similar to~\cite{lin2021efficient, neff2021donerf}.
%
%
With surface-guided rendering, the computational cost of synthesizing full-resolution images is significantly reduced.

\label{sgt}
During training, the geometry network $f_{\theta}$ is fixed.
The appearance network is optimized based on perceptual and adversarial losses.
%
We first use the learned semantic field to render multi-view semantic masks which are then translated into images using SPADE.
%
%
The perceptual loss~\cite{johnson2016perceptual} is adopted to compare the rendered image $\mathbf{C}$ and generated image $\mathbf{\hat{C}}$:
\begin{equation}
\mathcal{L}_{\text {feat }}(\mathbf{\hat{C}}, \mathbf{C})=\left\|\phi(\mathbf{\hat{C}})-\phi(\mathbf{C})\right\|_2^2,
\end{equation}
%
where $\phi$ denotes the VGG network~\cite{simonyan2014very}.
Perceptual loss makes training procedure more stable and faster.

Additionally, the adversarial loss is applied to make the rendered images more photorealistic and prevent blurriness caused by the inconsistency of input views, which is demonstrated in GANcraft~\cite{hao2021gancraft}.
Our rendered images are taken as ``fake'' samples, and generated images are taken as ``real'' samples.
We adopt the OASIS discriminator~\cite{sushko2020you} as our discriminator, and use the same generator and discriminator loss as~\cite{sushko2020you}.



%% file: fig/warpback.tex
\begin{figure}[t]
\centering
\includegraphics[width=1\linewidth]{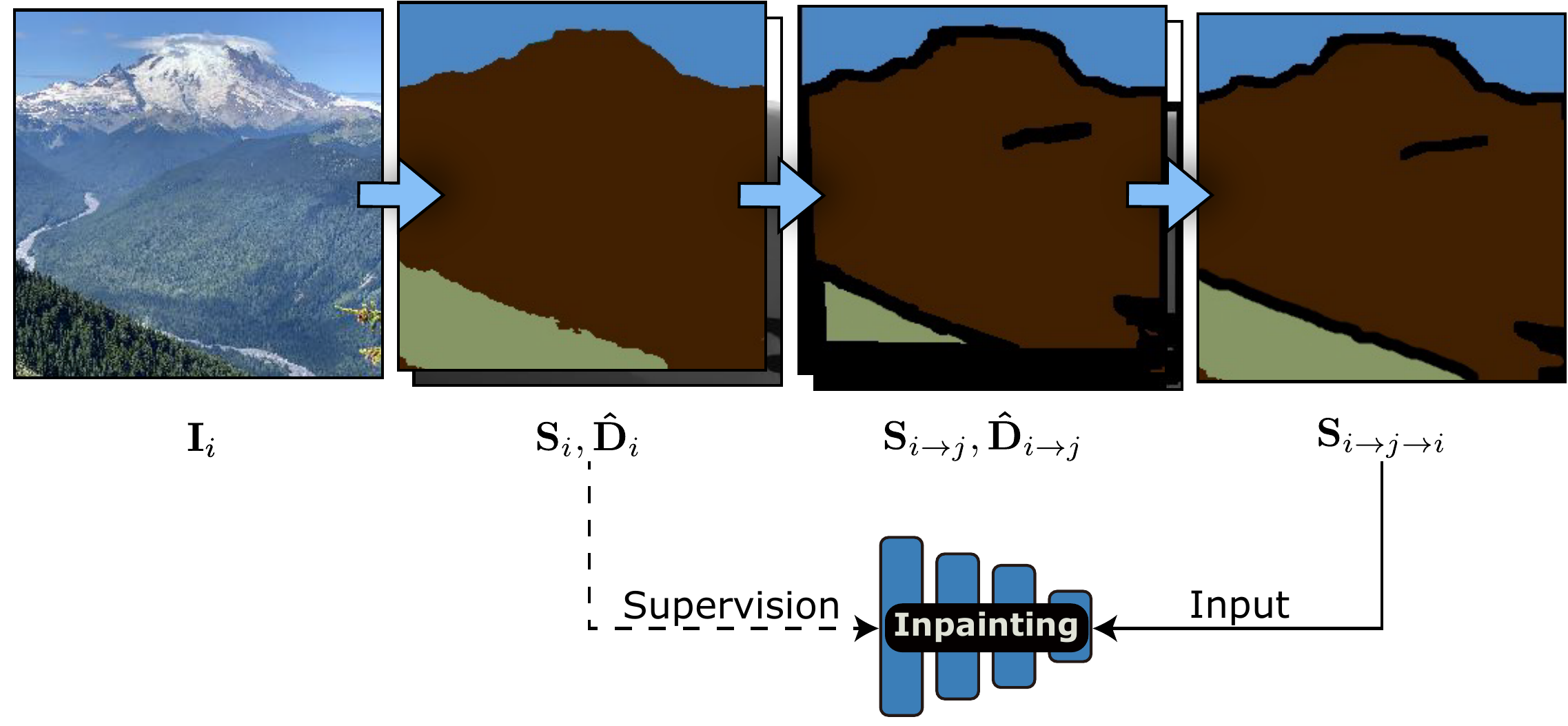}
\vspace{-1em}
\captionsetup{font={normalsize}} 
\caption{\textbf{Training a semantic inpainting network.} Our semantic inpainting network takes the $\mathbf{S}_{i\rightarrow j\rightarrow i}$
 as input and is trained to recover the $\mathbf{S}_i$.}
\label{fig:warpback}
\end{figure}

%% file: fig/vis_fuse.tex
\begin{figure}
\begin{center}
\setlength\tabcolsep{0.2em}
\newcommand{\mywidth}{0.112 \textwidth}

\begin{tabular}{cc|cc}

\includegraphics[width=\mywidth]{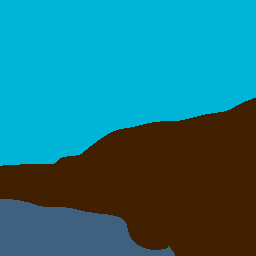}
& \includegraphics[width=\mywidth]{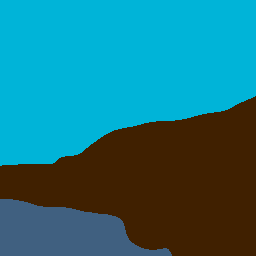}
&\includegraphics[width=\mywidth]{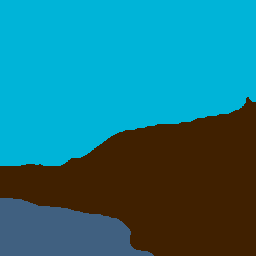}
&\includegraphics[width=\mywidth]{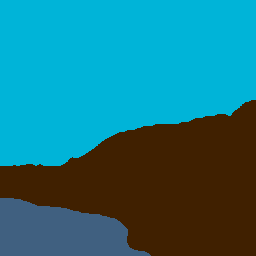}
\\
\includegraphics[width=\mywidth]{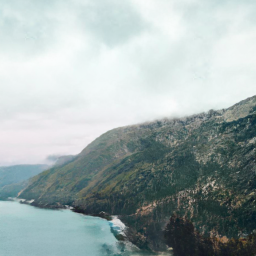}
& \includegraphics[width=\mywidth]{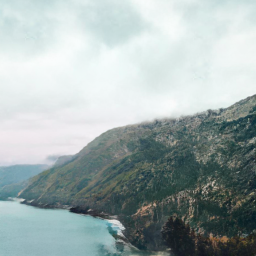}
&\includegraphics[width=\mywidth]{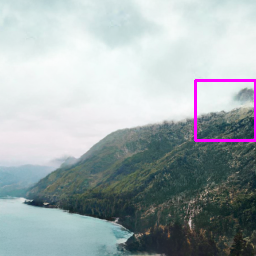}
&\includegraphics[width=\mywidth]{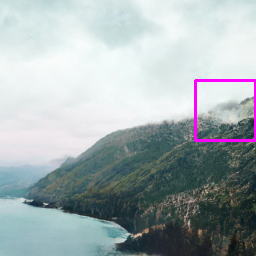}
\\
\includegraphics[width=\mywidth]{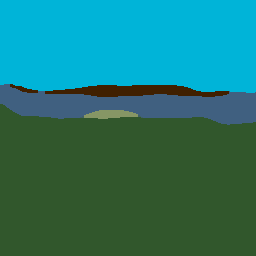}
& \includegraphics[width=\mywidth]{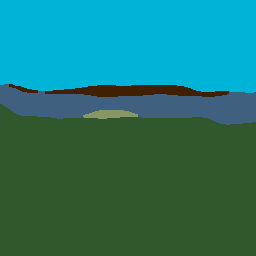}
&\includegraphics[width=\mywidth]{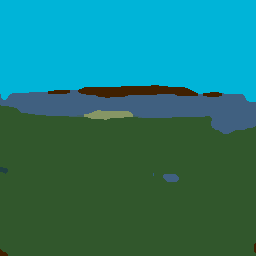}
& \includegraphics[width=\mywidth]{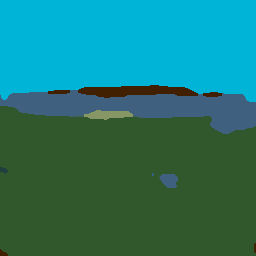}
\\

\includegraphics[width=\mywidth]{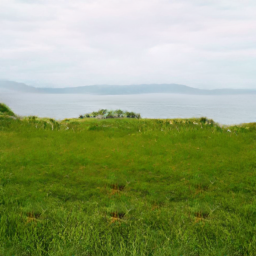}
&\includegraphics[width=\mywidth]{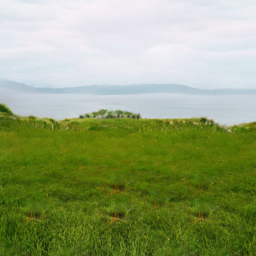}
&\includegraphics[width=\mywidth]{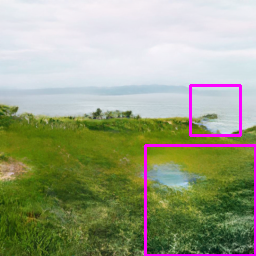}
&\includegraphics[width=\mywidth]{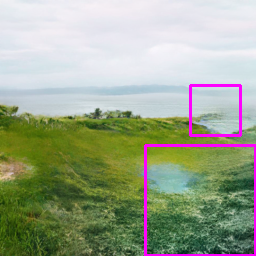}
\\
\multicolumn{2}{c}{w/ semantic field}
&
\multicolumn{2}{c}{w/o semantic field}
\end{tabular}
\end{center}
\captionsetup{font={normalsize}} 
\caption{\textbf{The effectiveness of semantic field.} The cropped patch clearly indicates the minor change in the semantic masks across different viewpoints (the first and second columns are adjacent viewpoints) brings the unwanted large region change in RGB images generated by SPADE.}

\label{fig:vis_fuse}
\end{figure}

%% file: sec_4_exp.tex
\input{exp/fid_kid.tex}

\section{Experiments}
\subsection{Datasets}
We use the LHQ dataset~\cite{skorokhodov2021aligning} to train our SPADE and semantic refinement network. The LHQ dataset is a large collection of landscape photos collected from the Internet.
To prepare the semantic label maps for each image, we use  COCO-Stuff~\cite{caesar2018coco} to train a DeepLab v2~\cite{chen2017deeplab} network.
The training data for semantic inpainting is synthesized by the back-warping strategy as mentioned in Section~\ref{sr}. 

\subsection{Implementation details}
To train the semantic field, 1024 rays are sampled per learning iteration. For the neural appearance field, we train the neural appearance field at $256\times 256$ resolution.
We train our semantic field and appearance field on 1 NVIDIA RTX 3090 with 24GB of memory.
%
For the semantic field, each model is trained for 48k iterations with batch size 1, which takes approximately 13 hours.
For the appearance field, each model is trained for 12k iterations with batch size 4, which takes approximately 3 hours.
We use a combination of GAN loss, L2 loss, and perceptual loss for the appearance field. 
Their weights are 1.0, 10.0, and 10.0, respectively. More training details are described in the supplementary material.

\subsection{Metrics}
Following Gancraft~\cite{hao2021gancraft}, we use both quantitative and qualitative metrics to evaluate our synthesis methods.

\paragraph{Quantitative metrics.}
We adopt the widely used metrics, FID~\cite{heusel2017gans, parmar2021cleanfid} and KID~\cite{binkowski2018demystifying, parmar2021cleanfid}, to measure the distance between the real and generated distributions.
%
Our experiments are conducted on 6 test scenes. A collection of landscape images from Flickr is obtained to evaluate the quality of our generated images. We make sure that test images are not presented in the LHQ dataset for training. The input semantic mask is obtained by a pretrained semantic segmentation model~\cite{chen2017deeplab}.
%
%
%
%
For each test scene, we render 330 images using a randomly sampled style code from uniformly sampled camera poses. 
For both FID and KID
metrics, lower values indicate better image quality.

\paragraph{Qualitative metrics.}
To quantify the aspects that are not addressed by automatic evaluation metrics, we conduct a user study to compare our method with the baseline methods. Specifically, we consider two aspects for human observers: 1) view consistency and 2) photo-realism. 18 visual designers in a 3D content designing company are asked to assign a score on a continuous scale of 1-5 for each aspect per video, where 1 is worse, and 5 is best. Two video sequences of the same scene rendered by two different methods are presented at the same time in random orders. A total of 9 videos per method is rendered for evaluation for each user. For more details about the user study, please refer to the supplementary material.
\input{fig/exp.tex}

\subsection{Comparisons with baseline methods}

We compare our approach with four strong baselines. Each method evaluates at $256\times 256$ resolution.

\textbf{SPADE~\cite{Park2019SemanticIS}+3DPhoto~\cite{Shih20203DPU}.} 
Using layered depth images, 3DPhoto~\cite{Shih20203DPU} generates free-viewpoint videos from a single image, and it can be trained on the in-the-wild dataset. 
Following~\cite{Li2022InfiniteNatureZeroLP,Rombach2021GeometryFreeVS,Liu2021InfiniteNP} that directly use official pretrained models for evaluation, we also use the official pretrained models for our evaluation.
We combine 3DPhoto and SPADE by synthesizing a color image from the semantic mask using SPADE, then applying 3DPhoto to generate a novel view image. 


\textbf{SPADE~\cite{Park2019SemanticIS}+AdaMPI~\cite{han2022single}.}
AdaMPI~\cite{han2022single} regresses the MPI representation from an image with a network trained on the in-the-wild dataset. 
 We also use official pretrained models for our evaluation.
This method can also be combined with SPADE to apply in this setting. 
We obtain generated images from the given semantic mask using SPADE 
and then use AdaMPI to synthesize the multi-plane images.

\textbf{SPADE~\cite{Park2019SemanticIS}+InfiniteNatureZero~\cite{Li2022InfiniteNatureZeroLP}.}
InfiniteNature-Zero is a model that generates flythrough videos of natural scenes, beginning with a single image. It is trained on the LHQ dataset.   We obtain generated images from the given semantic mask using SPADE 
and then use InfiniteNature-Zero to perform perceptual view generation according to the camera poses used in our test set. 

\textbf{$\text{GVS}^\ast$~\cite{Habtegebrial2020GenerativeVS}.}
We carefully train GVS~\cite{Habtegebrial2020GenerativeVS} on the LHQ dataset, 
using the same training strategy as AdaMPI~\cite{Shih20203DPU}, which takes approximately 2 days.
To ensure its style is consistent with other methods, it is finetuned on the test scenes.
We find the official inpainting network pretrained by AdaMPI~\cite{Shih20203DPU} can not deal with large disocclusion regions.
To prevent these regions from affecting GVS, GVS is not supervised in these areas.
GVS is trained on 3 NVIDIA TITAN A100 40GB graphics cards with 16 batch sizes. 



\figref{fig:exp} shows the results of generated videos of different methods. 
Directly combining SPADE with 3DPhoto or AdaMPI does not fully utilize the semantic layout information. 
Therefore, they are prone to erroneously inpainting disocclusion regions. In addition, 
our approach can render free-viewpoint videos with a wide range of viewpoints, because it is rather easy for our semantic inpainting network to synthesize a large disocclusion region and the appearance information at these regions can be generated via SPADE effortlessly.
As $\text{GVS}^\ast$ only produces multiple planes at fixed depths, 
it struggles to represent complex scene layouts for natural images. 
InfiniteNatureZero~\cite{Li2022InfiniteNatureZeroLP} is not designed for novel view synthesis and struggles to produce realistic results for the input camera trajectory is different from the training.  Besides, InfiniteNatureZero and  $\text{GVS}^\ast$ uses a 2D CNN to generate or refine RGB images, which do not guarantee the inter-view consistency. 
On the contrary, our method constructs a continuous neural field to fuse appearance information from SPADE, so that more realistic and consistent results can be obtained.

\input{exp/user_study}
As shown in Table~\ref{table:fid_kid}, our approach outperforms existing baselines, achieving the smallest FID and KID.
Furthermore, Table~\ref{table:user_study} indicates that users prefer our method and rate our videos the most view-consistent and realistic compared to others. 

%% file: exp/fid_kid.tex
\begin{table}
\centering
\begin{tabular}{ccc}
\hline
Methods & FID$\downarrow$ & KID$\downarrow$\\
\hline
SPADE~\cite{Park2019SemanticIS}+InfiniteNatureZero~\cite{Li2022InfiniteNatureZeroLP} & 149.80 & 0.080\\
SPADE~\cite{Park2019SemanticIS}+3DPhoto~\cite{Shih20203DPU} & 127.74 & 0.064 \\
SPADE~\cite{Park2019SemanticIS}+AdaMPI~\cite{han2022single} & 185.96 & 0.115\\
$\text{GVS}^\ast$~\cite{Habtegebrial2020GenerativeVS} & 141.64 & 0.084\\

Ours     & \textbf{109.85} & \textbf{0.050} \\
\hline
\end{tabular}

\captionsetup{font={normalsize}} 
\caption{\textbf{Quantitative comparisons on the LHQ dataset.} ``SPADE + *'' means a two-stage pipeline that first generates an image with SPADE and then performs single-view view synthesis. ``GVS$^\ast$'' means that we train GVS on the LHD dataset using the strategy in AdaMPI.}
\label{table:fid_kid}
\end{table}

%% file: fig/exp.tex
\begin{figure*}
\begin{center}
\setlength\tabcolsep{0.2em}
\newcommand{\mywidth}{0.15 \textwidth}

\begin{tabular}{cccccccc}

\includegraphics[width=\mywidth]{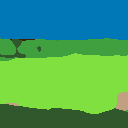}
& \includegraphics[width=\mywidth]{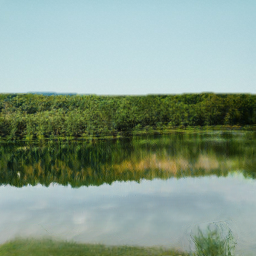}
&\includegraphics[width=\mywidth]{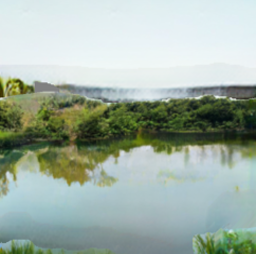}
&\includegraphics[width=\mywidth]{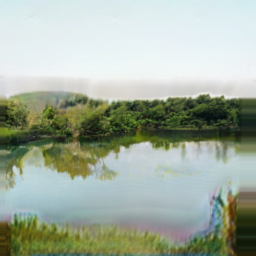}
&\includegraphics[width=\mywidth]{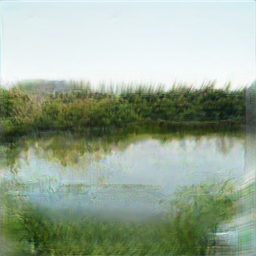}
&\includegraphics[width=\mywidth]{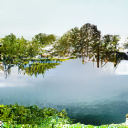}
\\
\includegraphics[width=\mywidth]{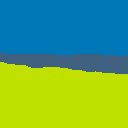}
& \includegraphics[width=\mywidth]{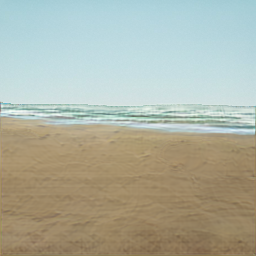}
&\includegraphics[width=\mywidth]{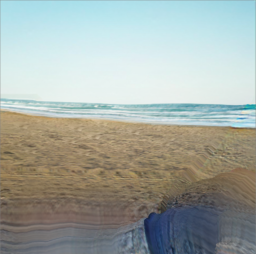}
&\includegraphics[width=\mywidth]{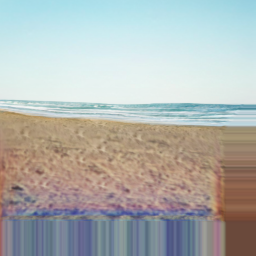}
&\includegraphics[width=\mywidth]{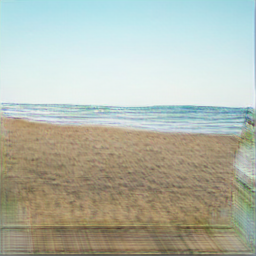}
&\includegraphics[width=\mywidth]{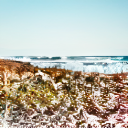}
\\
\includegraphics[width=\mywidth]{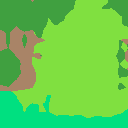}
& \includegraphics[width=\mywidth]{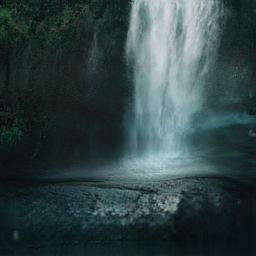}
&\includegraphics[width=\mywidth]{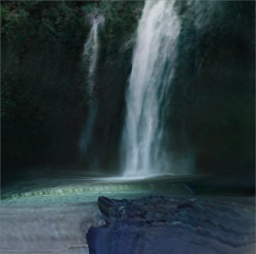}
&\includegraphics[width=\mywidth]{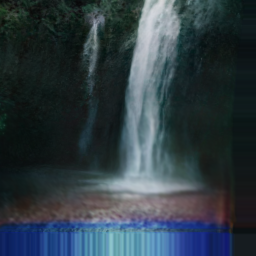}
&\includegraphics[width=\mywidth]{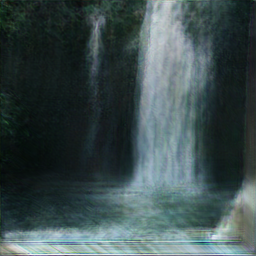}
&\includegraphics[width=\mywidth]{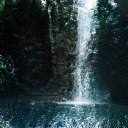}

\\
\includegraphics[width=\mywidth]{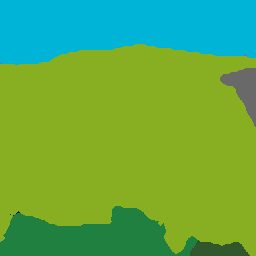}
& \includegraphics[width=\mywidth]{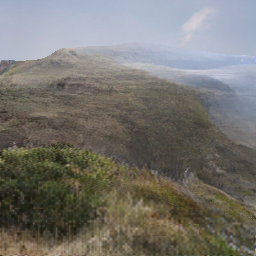}
&\includegraphics[width=\mywidth]{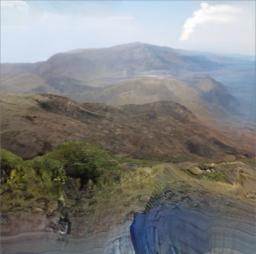}
&\includegraphics[width=\mywidth]{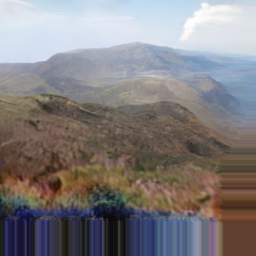}
&\includegraphics[width=\mywidth]{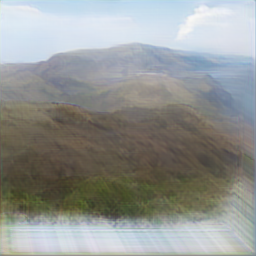}
&\includegraphics[width=\mywidth]{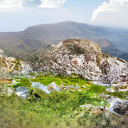}

\\
\includegraphics[width=\mywidth]{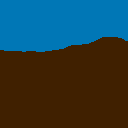}
& \includegraphics[width=\mywidth]{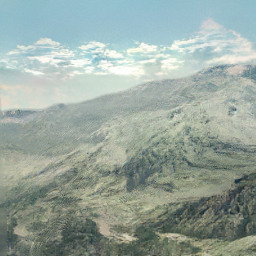}
&\includegraphics[width=\mywidth]{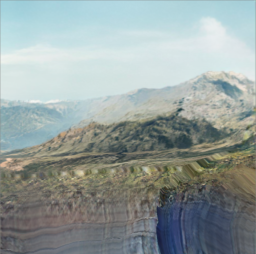}
&\includegraphics[width=\mywidth]{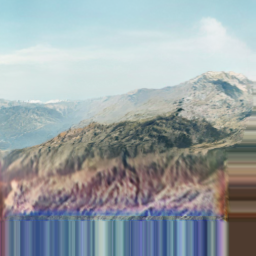}
&\includegraphics[width=\mywidth]{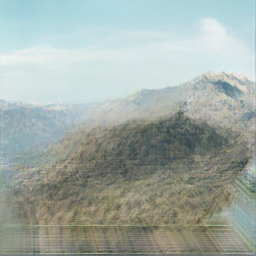}
&\includegraphics[width=\mywidth]{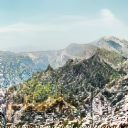}

\\
Semantic Mask
&Ours
& 3DPhoto~\cite{Shih20203DPU}
& AdaMPI~\cite{han2022single}
& $\text{GVS}^\ast$~\cite{Habtegebrial2020GenerativeVS}
& 
\footnotesize
InfiniteNatureZero~\cite{Li2022InfiniteNatureZeroLP}


\end{tabular}

\end{center}
\vspace{-0.2em}
\captionsetup{font={normalsize}} 
\caption{\textbf{Qualitative comparisons on the LHQ dataset.} We produce more realistic rendering compared to all baselines, which are demonstrated by the supplementary video.}
\vspace{-0.2em}
\label{fig:exp}
\end{figure*}

%% file: exp/user_study.tex
\begin{table}
\centering
\resizebox{0.45\textwidth}{!}{
\begin{tabular}{ccc}
\hline
Methods & Consistency$\uparrow$ & Realism$\uparrow$\\
\hline

SPADE~\cite{Park2019SemanticIS}+InfiniteNatureZero~\cite{Li2022InfiniteNatureZeroLP} & 1.31 & 2.06\\
SPADE~\cite{Park2019SemanticIS}+AdaMPI~\cite{han2022single}     &3.63   & 1.75 \\
SPADE~\cite{Park2019SemanticIS}+3DPhoto~\cite{Shih20203DPU}      &3.94 & 2.19 \\
$\text{GVS}^\ast$~\cite{Habtegebrial2020GenerativeVS}   &2.63 &  2.13 \\
Ours & \textbf{4.11} & \textbf{3.13} \\

\hline
\end{tabular}
}
\captionsetup{font={normalsize}} 
\caption{\textbf{Human preference scores.} Our method achieves the highest photo-realism and multi-view consistency according to human raters. }
\label{table:user_study}
\end{table}

%% file: sec_5_ablation.tex
\subsection{Ablation studies}
\label{ablation}
\input{fig/ablation.tex}
We conduct ablation studies to demonstrate the importance of each component of our method on one test scene.
Our key idea is decomposing the task of view synthesis from a semantic mask into two simpler steps, which first generate multi-view semantic masks and then produce RGB images with SPADE for learning a neural scene representation.
To demonstrate the effectiveness of this idea, we design a naive baseline where we use a pre-trained RGB inpainting network to generate multi-view images for recovering the 3D scene.
All methods use the same depth map for a fair comparison.
Specifically, We generate RGB images by SPADE from the given semantic mask and then use a monocular depth estimation model to predict the depth maps of the generated image.
The generated image is then warped to novel views. 
To infill disocclusion regions, we apply a pretrained RGB inpainting network to images at novel views, which output the final multi-view RGBD images.
The pretrained RGB and depth inpainting networks are the official models from AdaMPI \cite{han2022single}.
We abbreviate this baseline as ``RGB inpainting''.
As shown in \figref{fig:ablation}, this model fails to produce photorealistic results under big viewpoint changes.
This is because it is challenging for an RGB inpainting network trained on single-view datasets to infill such large missing areas.
In contrast, our approach renders high-quality images, which indicates that semantic masks are easier to inpaint than RGB images.

The second important design is our semantic field fusion module, which leverages the semantic field to denoise and fuse semantic masks generated by the inpainting network.
To illustrate the necessity of this module, we design the baseline where the infilled semantic masks are directly fed to SPADE to produce RGB images for learning the scene representation. For this baseline, we use the same geometry as our full method from the semantic field. We abbreviate this baseline as ``ours w/o SF''.
The result in \figref{fig:ablation} indicates that although this model can synthesize reasonable contents in disocclusion regions, the rendered images tend to be blurry.
The reason is that the infilled semantic masks are not view-consistent, especially near semantic edges, resulting in that images generated by SPADE differing significantly across different viewpoints, which makes us difficult to reconstruct the 3D scene. Besides, to demonstrate that our generated multi-view semantic masks are better than those generated by GVS~\cite{Habtegebrial2020GenerativeVS}, we compare 
our method with it on the
scene used in ablation studies. Following ~\cite{qiao2022learning}, we utilize the average Negative Log-Likelihood (NLL) score to measure  the quality of generated semantic masks and the View Semantic Consistency  (VSC)  score to evaluate the consistency of generated semantic masks. Tab.~\ref{table:layout} shows that the quality and consistency of our generated semantic masks outperform those of GVS.
\input{exp/ablation2.tex}


%% file: fig/ablation.tex
\begin{figure}
\begin{center}
\setlength\tabcolsep{0.1em}
\newcommand{\mywidth}{0.112 \textwidth}

\begin{tabular}{cccccc}

&\includegraphics[width=\mywidth]{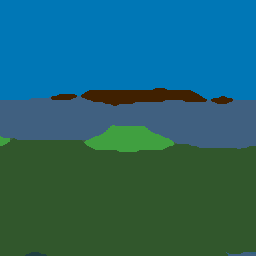}
&\includegraphics[width=\mywidth]{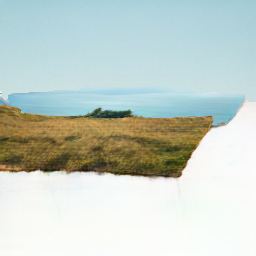}
&\includegraphics[width=\mywidth]{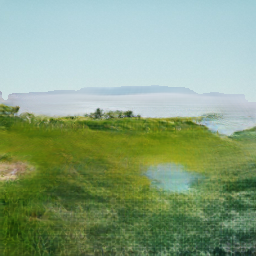}
&\includegraphics[width=\mywidth]{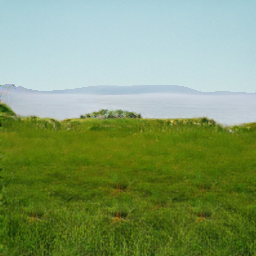}
\\
&  \scriptsize{Input semantic mask}
&  \scriptsize{RGB inpainting}
&  \scriptsize{Ours w/o SF}
&  \scriptsize{Ours}


\end{tabular}

\end{center}
\captionsetup{font={normalsize}} 
\vspace{-0.4em}

\caption{\textbf{Qualitative results of ablation studies.} ``RGB inpainting'' denotes the neural scene representation learned on multi-view images generated by an image inpainting network. 
This model fails to produce plausible image contents in disocclusion regions. ``Ours w/o SF'' denotes the scene representation learned on images generated from the (post-inpainting) semantic masks and has the same geometry as our full method. The inconsistency of semantic masks causes large changes in RGB images, resulting in degraded rendering quality. Please see the supplementary video for more visual comparisons.}
\vspace{-0.4em}

\label{fig:ablation}
\end{figure}

%% file: exp/ablation2.tex
\begin{table}
\centering
\begin{tabular}{ccc}
\hline
\centering

Methods &  NLL$\downarrow$ & VSC$\downarrow$\\
\hline
Ours & 1.60 & 0.049 \\
GVS &  3.33 & 0.089\\
\hline

\end{tabular}
\vspace{-0.2em}

\captionsetup{font={normalsize}} 
\caption{\textbf{Comparison with GVS.} Our mulit-view semantic masks are more consistent and have better quality.}

\label{table:layout}
\end{table}

%% file: sec_6_conclusion.tex
\section{Conclusion}

In this work, we expanded the AI-enabled content creation tool by using a single semantic mask to produce a 3D scene that can be rendered from arbitrary viewpoints. 
Our method requires single-view image collections for training, without the need for multi-view images.
To achieve this, we proposed a novel pipeline that first learns an inpainting network to generate novel views of the input semantic mask and then optimizes a 3D semantic field to render view-consistent semantic masks, which are subsequently fed into a 2D generator SPADE to produce RGB images for learning a neural scene representation.
Experiments demonstrated that our method can produce satisfactory photorealistic and view-consistent results that significantly outperform baseline methods. 
We believe that, like AI painting has changed the field, our method has great potential to revolutionize 3D content creation for humans.

\paragraph{Limitation.}
Our neural scene representation is optimized per scene, which takes a lot of time.
Amortized inference models are much more 
convenient for users to edit digital content with quick feedback.
How to train amortized inference models for view synthesis of natural scenes from a single semantic mask on the single-view datasets remains unexplored.
This problem is left for future work.

%% file: Supp_arxiv.tex
\appendix

\section{Implementation Details}

\subsection{Network architecture}
\paragraph{Semantic field.}
\figref{fig:mlp} shows our network architecture of the semantic field. 
Our network takes the 3D point $\mathbf{x}$ as the input and outputs SDF $d$, semantic logits $\mathbf{s}$.
For the SDF field, we follow the official implementation of ManhattanSDF~\cite{guo2022manhattan}.
Following NeRF~\cite{mildenhall2021nerf}, we utilize a positional encoding to map the $\mathbf{x}$ to a higher dimensional feature space: 
\begin{equation}
    \begin{aligned}
        &\gamma(\mathbf{x})=\left[\gamma_0(\mathbf{x}), \gamma_1(\mathbf{x}), \ldots, \gamma_{L-1}(\mathbf{x})\right]\\
        &\gamma_j(\mathbf{x})=\left[\sin \left(2^j \pi \mathbf{x}\right), \cos \left(2^j \pi \mathbf{x}\right)\right]
    \end{aligned}
\end{equation}
where $L = 6$.

\paragraph{Appearance field.}

\figref{fig:generator} and \figref{fig:generator_mlp} illustrate the network architecture of our appearance field.
The network architecture of the feature planes generator follows the official implementation of FastGAN~\cite{liu2020towards}.
The generator takes a 256-length noise vector as input and outputs a $512 \times 512 \times 180$ feature map, 
which is then split to form three 60-channel feature planes. 
The 256-length noise vector is fixed. 

\paragraph{Semantic inpainting network.}

Following~\cite{Liu2021InfiniteNP,Li2022InfiniteNatureZeroLP}, 
we view our semantic inpainting step as an image-to-image translation task
and adopt the SPADE generator~\cite{Park2019SemanticIS} as our semantic inpainting network.

\paragraph{Label-conditional discriminator.}

Our discriminator is the same as the official implementation of OASIS~\cite{sushko2020you}.
\paragraph{2D sky generator.}
We directly use the generator of FastGAN~\cite{liu2020towards} to the produce the sky image plane at $512 \times 512$ resolution.
The padding mode of our sky image plane is ``reflection''.

\subsection{Loss function}
\paragraph{Semantic field.}
To optimize the semantic field, we use the following overall loss function:
\begin{equation}
\begin{aligned}
\mathcal{L}=&\lambda_0 \mathcal{L_{\text{depth}}}+ \lambda_1 \mathcal{L}_{\text{trans}} + \lambda_2 \mathcal{L}_{\mathbf{P}} +\lambda_3 \mathcal{L}_{\text {eikonal }} \\ & +\lambda_4 \mathcal{L}_{\text {rank }}
+\lambda_5 \mathcal{L}_{\text {src}} ,
\end{aligned}
\end{equation}
where $\lambda_0 = 0.1, \lambda_1=10, \lambda_2=1,  \lambda_3=0.01, \lambda_4 =0.1, \lambda_5 = 1$.
The $\mathcal{L_{\text{depth}}}$, $\mathcal{L}_{\text{trans}}$ and $\mathcal{L}_{\mathbf{P}}$ are defined in the main paper.

Following common practice, an Eikonal loss~\cite{gropp2020implicit} is added to regularize the SDF field.

\begin{equation}
\mathcal{L}_{\text {eikonal}}=\sum_{\mathbf{x} \in \mathcal{X}}\left(\left\|\nabla f_\theta(\mathbf{x})\right\|_2-1\right)^2,
\end{equation}
where $\mathcal{X}$ denotes a set of near-surface and uniformly sampled points~\cite{yariv2021volume}.

To encourage the ordinal relation of rendered depth maps matches the ground-truth ordinal relation, we adopt ranking loss~\cite{chen2016single, xian2020structure}, as shown in Equation~\ref{rank}.
\begin{small}
\begin{equation}\label{rank}
\begin{aligned}
\mathcal{L}_{\text{rank}}= \begin{cases}\log \left(1+\exp \left(-\ell\left(p_0-p_1\right)\right)\right), & \ell \neq 0 \\ \left(p_0-p_1\right)^2, & \ell=0 ,\end{cases}
\end{aligned}
\end{equation}
where \end{small}$p_0=\mathbf{D}_{(u_0,v_0)}$ and $p_1=\mathbf{D}_{(u_1,v_1)}$ are the rendered depth values,
where $(u_0,v_0)$ and $(u_1,v_1)$  denote the locations of the randomly sampled pixels. 
$\ell$ is induced by the predicted monocular depth values and a tolerance threshold $\tau$. 
$p_0^*=\mathbf{\hat{D}}_{(u_0,v_0)}$, $p_1^*=\mathbf{\hat{D}}_{(u_1,v_1)}$ are the the predicted monocular depth values at location $(u_0,v_0)$, $(u_1,v_1)$:
\begin{small}
\begin{equation}\label{ol}
    \ell= \begin{cases}+1, & p_0^* - p_1^*>=\tau \\ -1, & p_0^* - p_1^*<=\tau \\ 0, & \text { otherwise }.\end{cases}
\end{equation}
\end{small}

In the main paper, we use a depth map $\mathbf{\hat{D}_\text{src}}$ generated by a monocular depth estimation network~\cite{ranftl2020towards} to warp the input semantic mask to novel views.
The depth map $\mathbf{\hat{D}_\text{src}}$ is also warped to novel views, yielding warped depth maps $\mathbf{\hat{D}_\text{src\{1...N\}}}$  at novel views $\{1...N\}$.
To further utilize the geometry information of the depth map $\mathbf{\hat{D}_\text{src}}$, we encourage the rendered depth map $\mathbf{D_\text{src\{1...N\}}}$ to equal  the depth map $\mathbf{\hat{D}_\text{src\{1...N\}}}$ at each novel view $\{1...N\}$,
 using Equation~\ref{src}:
\begin{small}
\begin{equation}\label{src}
    \mathcal{L}_{src}= \sum^{N}_{i=1}|\mathbf{D_\text{src\{i\}}} - \mathbf{\hat{D}_\text{src\{i\}}}|.
\end{equation}
\end{small}
\input{fig/mlp}

\paragraph{Appearance field.}
To learn the appearance field, we use GAN loss, L2 loss, and perceptual loss. 
Their weights are 1.0, 10.0, and 10.0, respectively.
\paragraph{Semantic inpainting network.}
Cross-entropy loss is used to supervise the semantic inpainting network:

\begin{equation}
    \mathcal{L}=\sum_{\mathbf{p} \in \mathcal{P}} CE( \mathbf{P}^*(\mathbf{p}),  \mathbf{\hat{P}}(\mathbf{p})),
\end{equation}
where $\mathbf{P}^*(\mathbf{p})$ is the output of the semantic inpainting network, $\mathbf{\hat{P}}(\mathbf{p})$ denotes the ground-truth label, and $\mathcal{P}$ means the set of image pixels. 
\input{fig/supp_video.tex}

\subsection{Other details}

\paragraph{Semantic field.}
We use the sampling algorithm proposed in VolSDF~\cite{yariv2021volume} and choose Adam~\cite{kingma2014adam} as our optimizer, 
with $\beta_1 = 0.9$, $\beta_2 = 0.999$, and a learning rate of $5\times 10^{-4}$.

\paragraph{Appearance field.}
We adopt Adam~\cite{kingma2014adam} as our optimizer of generator, 
with $\beta_1 = 0.9$, $\beta_2 = 0.999$, and a learning rate of $5\times 10^{-4}$.
For discriminator, We also adopt Adam~\cite{kingma2014adam},
with $\beta_1 = 0.9$, $\beta_2 = 0.999$, and a learning rate of $4\times 10^{-4}$.
To prevent the discriminator from overfitting the training data, 
the differentiable augmentation technique~\cite{zhao2020differentiable}
 is utilized for the training of the discriminator and the generator.

\paragraph{SPADE.}
We adopt the official implementation and training procedure of SPADE~\cite{Park2019SemanticIS} and train it on the LHQ~\cite{skorokhodov2021aligning} dataset. 
Following common practice, the DeepLabv2 model~\cite{chen2017deeplab} pre-trained on the COCO-Stuff~\cite{caesar2018coco} is adopted to obtain a semantic mask for each image in the LHQ dataset.

\paragraph{Semantic inpainting network.}
We train the semantic inpainting network on the LHQ~\cite{skorokhodov2021aligning} dataset. 
The semantic inpainting network is trained on 2 NVIDIA RTX 3090 at  $128\times 128$ resolution with batch size 32. 
The network is trained for a total of 60K iterations, which takes approximately 2 days.
For test time, the input semantic masks are downsampled to the $128\times 128$ resolution and the output semantic masks are upsampled to $256\times 256$ resolution.

\input{fig/generator_mlp}
\input{fig/generator}

\section{Experiment Details}
\subsection{Evaluation settings}
To ensure the diversity of our test scenes, our test set contains diverse classes:
\textit{sky, sea, mountain, hill, sand, rock, waterfall, river, tree, snow, grass}.

We preprocess the depth maps used in our experiments via the code provided by 3DPhoto~\cite{Shih20203DPU}. 
All methods use the same depth maps and style codes for a fair comparison.
The camera pose for evaluation is uniformly sampled from a unit cube. 

\subsection{Discussion of user study}
We conducted the user study following the settings of \cite{cai2021unified,michel2022text2mesh,yao2022dfa} and used the metric of widely-used Mean Option Score (MOS) ranging from 1-5 \cite{seufert2019fundamental} to evaluate our algorithm. Users who participated in the study are designers who are familiar with 3D content.
Prior to the study, we give each participant a brief and one-to-one introduction to the concept of view consistency and photo-realism (being realistic as captured by real cameras). 
During the study session, we found almost 100\% users preferred our method's result, showing our proposed method's superior performance. 
The ratings they provide further illustrate the degree of superiority.
Although the 3DPhoto~\cite{Shih20203DPU} and AdaMPI~\cite{han2022single} guarantee view consistency theoretically,
they tend to generate unrealistic content at the views far from the center view, 
leading to the fact that users are more likely to lower their view-consistency scores. 
Our method produces realistic and consistently high-quality content with fewer artifacts. 

\subsection{More details of ablation studies}
The quantitative results of our ablation studies are shown in Table~\ref{table:ab}.
We render 240 images for evaluation.
Because the ablation studies are only conducted on a single test set, the generated distribution is really different from the real distribution,
thus causing our full model to obtain higher FID and KID than the quantitative results in the main paper. 
\input{exp/ablation.tex}

\section{Visualization}
\input{fig/spade.tex}

    Directly translating multi-view semantic masks obtained by our semantic field to RGB images through SPADE~\cite{Park2019SemanticIS} fails to produce multi-view consistent results, as shown in \figref{fig:spade}.
    We  abbreviate this pipeline as ``$\text{SPADE}^\ast$''. 
    \figref{fig:sv} shows several cases of test scenes. 
    For more visualization results, we recommend watching the supplemental video. 
    
\section{Results on Other Kinds of Scenes}
Our method is general
and can extend to other kinds of scenes, including indoor
scenes, as shown in \figref{fig:indoor}. In the main paper, we focus on
natural scenes, as they are well-suited to our objectives. We
will add more results on other types of scenes.

\input{fig/indoor.tex}

\section{Discussion of Our Method}
Despite the seemingly long
pipeline, our high-level idea is clear and simple, which is to
divide a difficult task into two simpler steps: first predicting
multi-view semantic masks as intermediate representations
and then translating semantic masks to RGB images. The
additional modules of learning neural fields are just used for ensuring multi-view consistency.



%% file: fig/mlp.tex
\begin{figure}[t]
\centering
\includegraphics[width=1\linewidth]{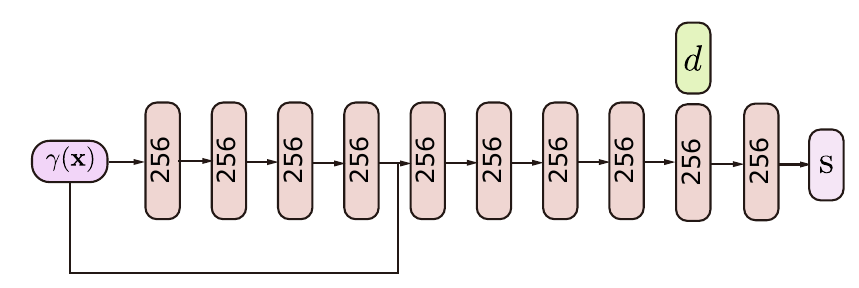}
\vspace{-1em}
\captionsetup{font={normalsize}} 
\caption{\textbf{Semantic field.} All layers are linear layers with
softplus activations except for the final layer.} 
\label{fig:mlp}
\end{figure}

%% file: fig/supp_video.tex
\begin{figure*}[t]
\begin{center}
\setlength\tabcolsep{0.2em}
\newcommand{\mywidth}{0.15 \textwidth}

\begin{tabular}{cccccc}

\includegraphics[width=\mywidth]{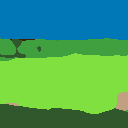}
& \includegraphics[width=\mywidth]{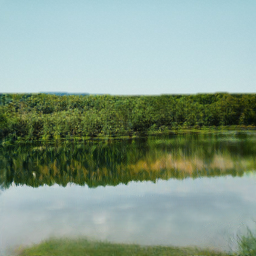}
&\includegraphics[width=\mywidth]{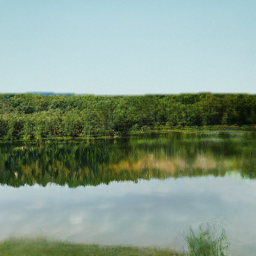}

&\includegraphics[width=\mywidth]{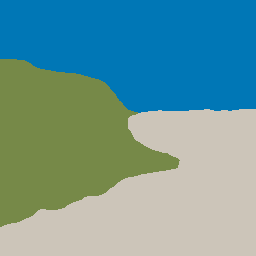}
& \includegraphics[width=\mywidth]{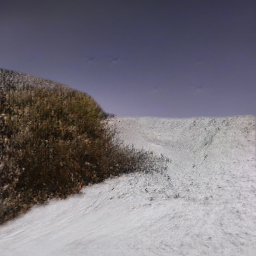}
&\includegraphics[width=\mywidth]{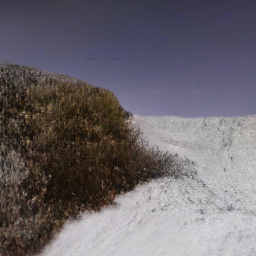}

\\
\includegraphics[width=\mywidth]{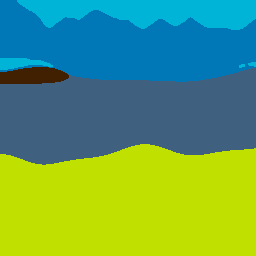}
& \includegraphics[width=\mywidth]{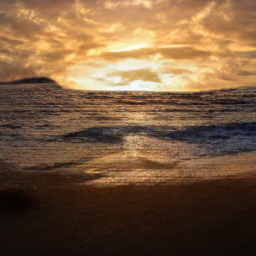}
&\includegraphics[width=\mywidth]{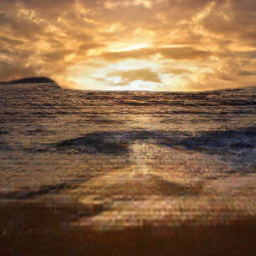}

&\includegraphics[width=\mywidth]{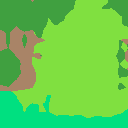}
& \includegraphics[width=\mywidth]{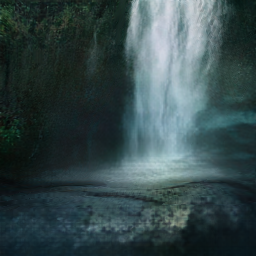}
&\includegraphics[width=\mywidth]{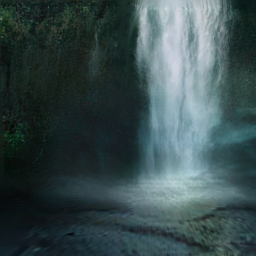}

\\
\includegraphics[width=\mywidth]{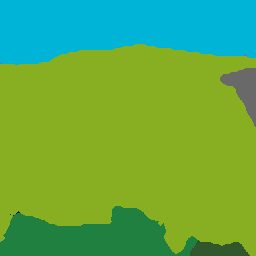}
& \includegraphics[width=\mywidth]{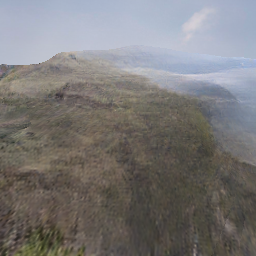}
&\includegraphics[width=\mywidth]{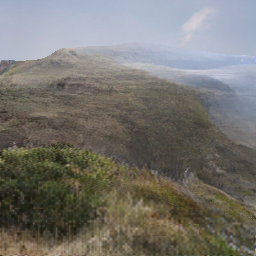}

&\includegraphics[width=\mywidth]{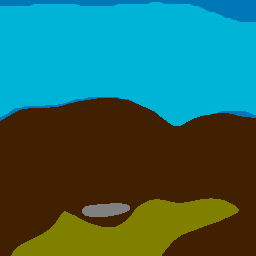}
& \includegraphics[width=\mywidth]{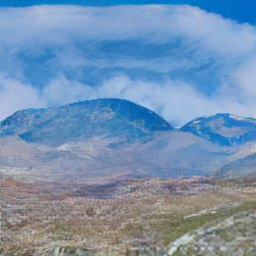}
&\includegraphics[width=\mywidth]{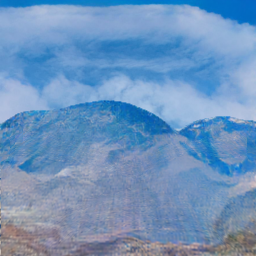}

\\
\includegraphics[width=\mywidth]{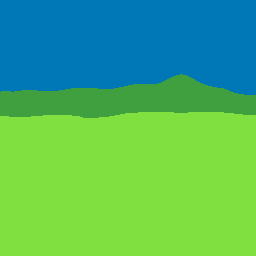}
& \includegraphics[width=\mywidth]{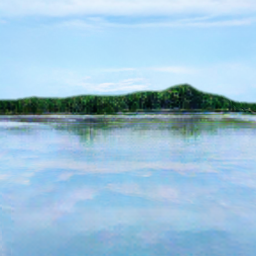}
&\includegraphics[width=\mywidth]{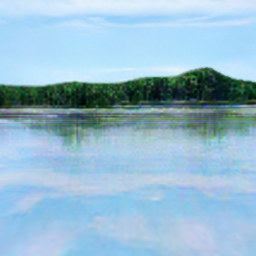}

&\includegraphics[width=\mywidth]{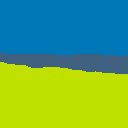}
& \includegraphics[width=\mywidth]{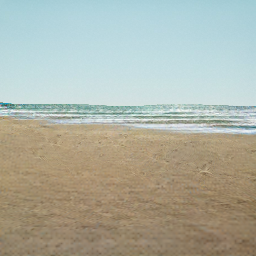}
&\includegraphics[width=\mywidth]{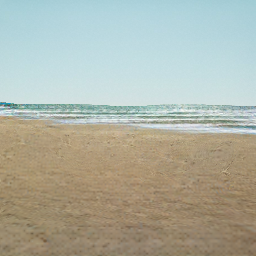}

\\
\small Semantic Mask
& \small Novel View 1
& \small Novel View 2
& \small Semantic Mask
& \small Novel View 1
& \small Novel View 2


\end{tabular}

\end{center}
\captionsetup{font={normalsize}} 
\caption{\textbf{Qualitative results.} Our results are  view-consistent and photorealistic, as demonstrated by the supplementary video.}
\label{fig:sv}
\end{figure*}

%% file: fig/generator_mlp.tex
\begin{figure}[t]
\centering
\includegraphics[width=0.7\linewidth]{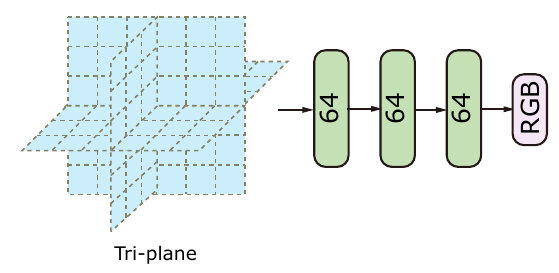}
\vspace{-1em}
\captionsetup{font={normalsize}} 
\caption{\textbf{The overview of the appearance field.} The feature retrieved from the tri-plane generator is converted to RGB value by an MLP with ReLU activations except for the final layer.} 
\label{fig:generator_mlp}
\end{figure}

%% file: fig/generator.tex
\begin{figure}[t]
\centering
\includegraphics[width=1\linewidth]{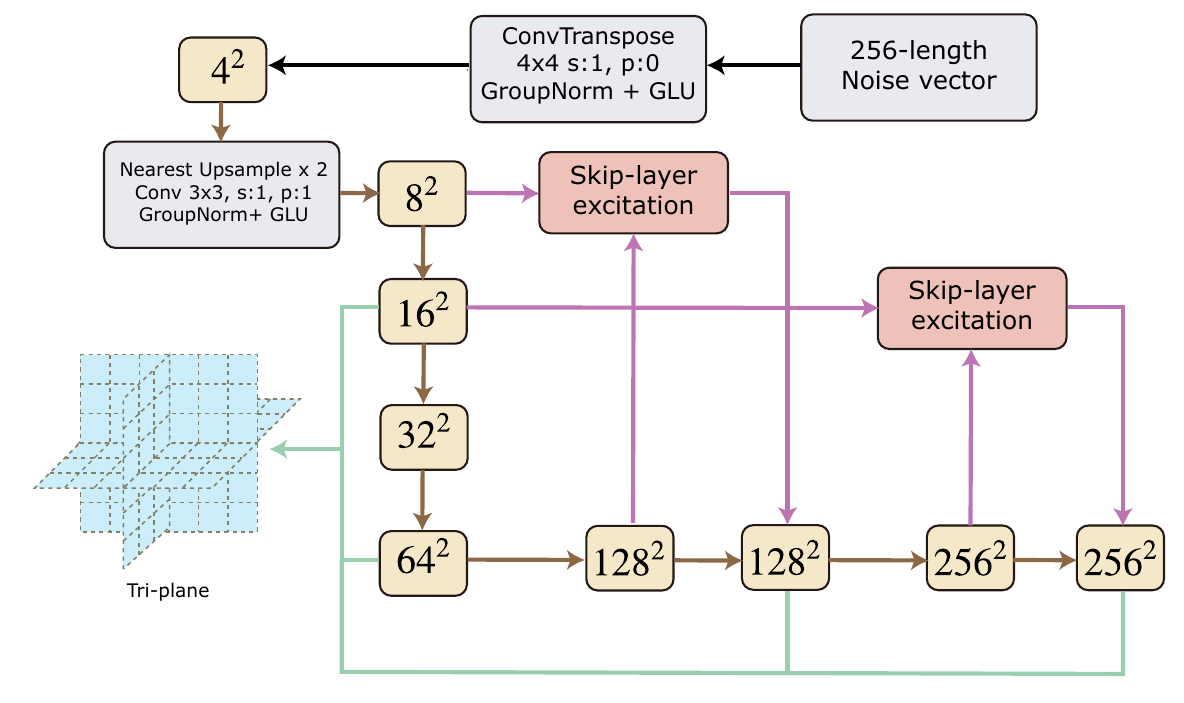}
\vspace{-1em}
\captionsetup{font={normalsize}} 
\caption{\textbf{The tri-plane generator.} The orange boxes denote feature maps with spatial size. 
The brown arrows denote an up-sampling module.
 The detailed architecture of skip-layer excitation and the channel number of each layer can be found in FastGAN~\cite{liu2020towards}. 
 The green arrows represent a $3\times 3$ 2D convolutions layer, which converts the intermediate feature maps to 45-channel feature maps. 
 The tri-plane feature map is formed by splitting and concatenating multi-resolution these 45-channel feature maps.} 
\label{fig:generator}
\end{figure}

%% file: exp/ablation.tex
\begin{table}
\centering
\begin{tabular}{ccc}
\hline
Methods & FID$\downarrow$ & KID$\downarrow$\\
\hline
RGB inpainting & 260.46 & 0.279 \\
Ours w/o SF & 250.70 & 0.236\\
Ours     & \textbf{166.94} & \textbf{0.124} \\
\hline
\end{tabular}

\captionsetup{font={normalsize}} 
\caption{\textbf{Quantitative results of our ablations.} ``RGB inpainting'' denotes the neural scene representation learned on multi-view images generated by an image inpainting network. 
``Ours w/o SF'' denotes the scene representation learned on images generated from the post-inpainting semantic masks. Our full model outperforms all ablated versions significantly, 
which demonstrates the effectiveness of our proposed framework.}
\label{table:ab}
\end{table}

%% file: fig/spade.tex
\begin{figure}[t]
\begin{center}
\setlength\tabcolsep{0.2em}
\newcommand{\mywidth}{0.112 \textwidth}

\begin{tabular}{cccc}

\includegraphics[width=\mywidth]{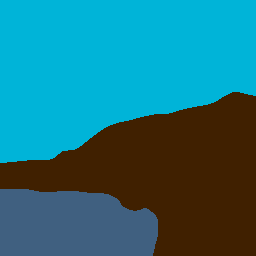}
& \includegraphics[width=\mywidth]{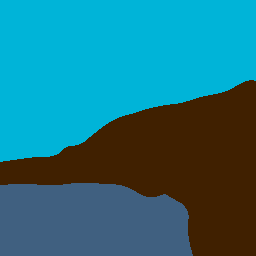}
&\includegraphics[width=\mywidth]{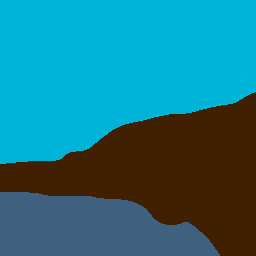}
&\includegraphics[width=\mywidth]{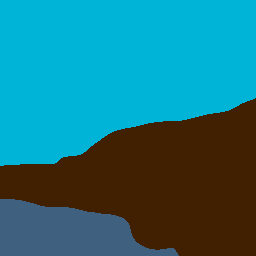}
\\
\includegraphics[width=\mywidth]{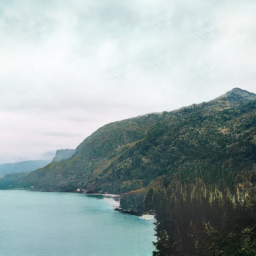}
& \includegraphics[width=\mywidth]{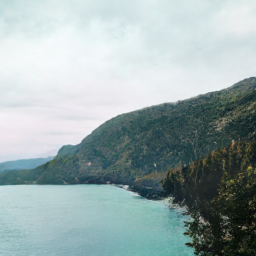}
&\includegraphics[width=\mywidth]{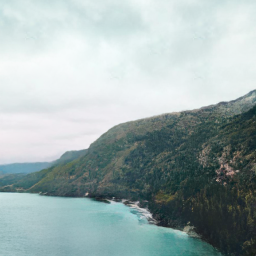}
&\includegraphics[width=\mywidth]{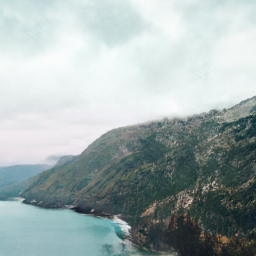}

\end{tabular}
\end{center}
\captionsetup{font={normalsize}} 
\caption{\textbf{The results of $\text{SPADE}^\ast$.} $\text{SPADE}^\ast$ denotes the pipeline which directly translates multi-view consistent semantic masks to RGB images via SPADE. 
Although the multi-view semantic masks are view consistent (first row), the generated RGB images via SPADE (second row) are inconsistent and have flickering artifacts between different views.}

\label{fig:spade}
\end{figure}

%% file: fig/indoor.tex
\begin{figure}[t]


\setlength\tabcolsep{0.1em}

\newcommand{\mywidth}{0.11 \textwidth}
\centering

\begin{tabular}{ccc}

\includegraphics[width=\mywidth]{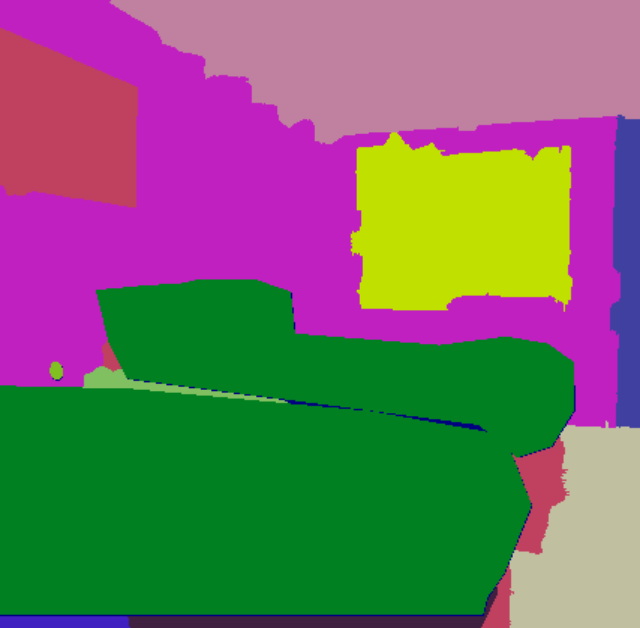}
&\includegraphics[width=\mywidth]{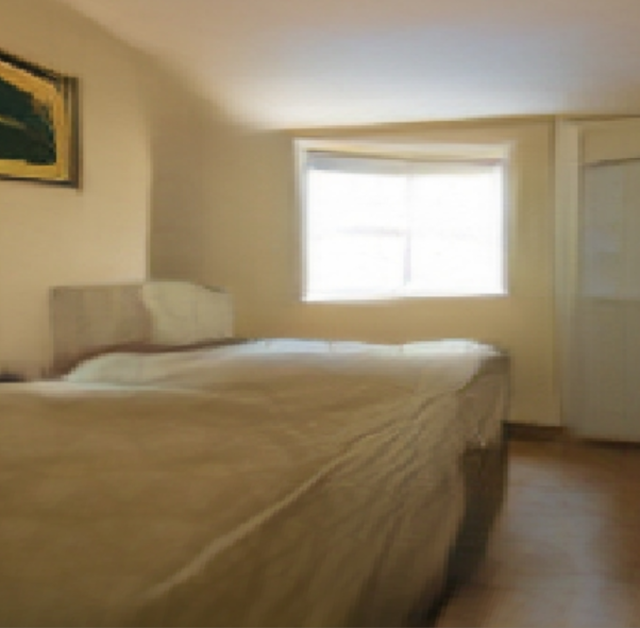}
& \includegraphics[width=\mywidth]{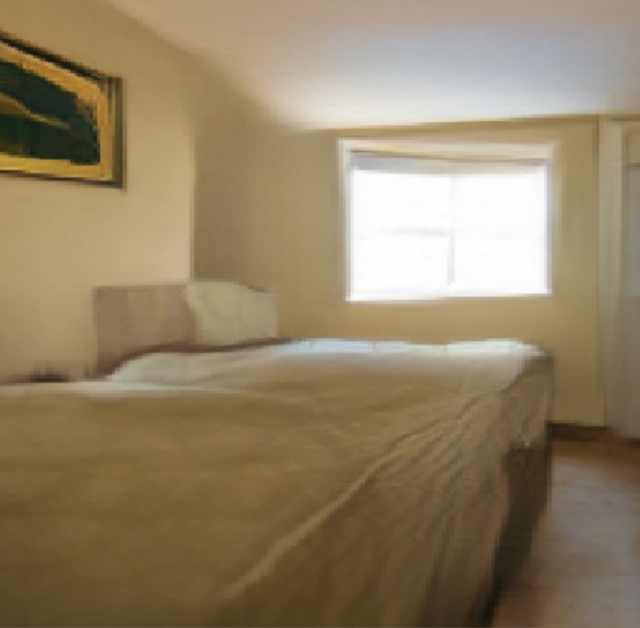}\\
\footnotesize
 Input mask
& \footnotesize View 1
& \footnotesize View 2

\end{tabular}

\captionsetup{font={normalsize}} 
\caption{\textbf{Results on indoor scenes.}}

\label{fig:indoor}








\end{figure}